\documentclass{article}

\usepackage{arxiv}
\usepackage{float} 
\usepackage[utf8]{inputenc} 
\usepackage[T1]{fontenc}    
\usepackage{hyperref}       
\usepackage{url}            
\usepackage{booktabs}       
\usepackage{amsfonts}       
\usepackage{nicefrac}       
\usepackage{microtype}      
\usepackage{lipsum}		
\usepackage{graphicx}
\usepackage{natbib}
\usepackage{doi}
\usepackage{tabularx} 
\usepackage{booktabs} 
\usepackage{algorithm}
\usepackage{algpseudocode}
\usepackage{amsmath}

\title{Advancing Data-driven Weather Forecasting: Time-Sliding Data Augmentation of ERA5}

\author{
    Minjong Cheon \\
    Center for Sustainable Environment Research\\
    Korea Institute of Science and Technology\\
    Seoul, South Korea \\
    \texttt{jmj2316@kist.re.kr} \\
    \And
    Daehyun Kang\thanks{Corresponding author.} \\
    Center for Sustainable Environment Research\\
    Korea Institute of Science and Technology\\
    Seoul, South Korea \\
    \texttt{dkang@kist.re.kr} \\
    \And
    Yo-Hwan Choi \\
    Center for Sustainable Environment Research\\
    Korea Institute of Science and Technology\\
    Seoul, South Korea \\
    \texttt{092935@kist.re.kr} \\
    \And
    Seon-Yu Kang \\
    Center for Sustainable Environment Research\\
    Korea Institute of Science and Technology\\
    Seoul, South Korea \\
    \texttt{sykang92@kist.re.kr} \\
}

\renewcommand{\shorttitle}{\textit{arXiv} Template}

\hypersetup{
pdftitle={A template for the arxiv style},
pdfsubject={q-bio.NC, q-bio.QM},
pdfauthor={Minjong},
pdfkeywords={First keyword, Second keyword, More},
}

\begin{document}
\maketitle

\begin{abstract}
Modern deep learning techniques, which mimic traditional numerical weather prediction (NWP) models and are derived from global atmospheric reanalysis data, have caused a significant revolution within a few years. In this new paradigm, our research introduces a novel strategy that deviates from the common dependence on high-resolution data, which is often constrained by computational resources, and instead utilizes low-resolution data (2.5 degrees) for global weather prediction and climate data analysis. Our main focus is evaluating data-driven weather prediction (DDWP) frameworks, specifically addressing sample size adequacy, structural improvements to the model, and the ability of climate data to represent current climatic trends. By using the Adaptive Fourier Neural Operator (AFNO) model via FourCastNet and a proposed time-sliding method to inflate the dataset of the ECMWF Reanalysis v5 (ERA5), this paper improves on conventional approaches by adding more variables and a novel approach to data augmentation and processing. Our findings reveal that despite the lower resolution, the proposed approach demonstrates considerable accuracy in predicting atmospheric conditions, effectively rivaling higher-resolution models. Furthermore, the study confirms the model's proficiency in reflecting current climate trends and its potential in predicting future climatic events, underscoring its utility in climate change strategies. This research marks a pivotal step in the realm of meteorological forecasting, showcasing the feasibility of lower-resolution data in producing reliable predictions and opening avenues for more accessible and inclusive climate modeling. The insights gleaned from this study not only contribute to the advancement of climate science but also lay the groundwork for future innovations in the field.
\end{abstract}


\section{Introduction}
In recent years, there has been a notable shift towards the adoption of deep learning-based models, driven by the remarkable capabilities of artificial intelligence. These models excel in the analysis of vast datasets, efficiently unveiling intricate underlying patterns \cite{dueben2018challenges}. This transition aligns with the growing trend of utilizing high-resolution data in climate and weather modeling, a transformation significantly propelled by the emergence of deep learning techniques \cite{bauer2023deep}. While numerous studies have substantiated the remarkable gains in forecast accuracy and granularity achieved through these methods, they come with a substantial set of challenges. The primary hurdle lies in the substantial computational GPU power and associated costs demanded by these models. Processing high-resolution data necessitates robust and resource-intensive computing infrastructures, often imposing prohibitive expenses. Consequently, the scalability and broad applicability of these sophisticated models are distinctly hampered by their reliance on high-resolution data sources \cite{guo2024fourcastnext}. Additionally, the data-driven weather prediction (DDWP) models rely on the training dataset in the past decades, which limits accuracy due to the lack of observational data and inconsistency among them due to climate change.

Our research explores an area that is relatively unexplored by utilizing low-resolution data, considering the limitations imposed by the dependence on high-resolution data. This study paves the potential way to further improve the efficiency and accuracy of the DDWP framework with a lower computing demand. This study aims to comprehensively assess the impact of sample size and model structural improvements on the performance of climate data analysis within the DDWP framework. Specifically, it will investigate:

\begin{enumerate}
    \item Data Sufficiency in DDWP: Evaluate whether the current sample size of training data is adequate for reliable climate analysis. This involves exploring the relationship between sample size and performance improvement.
    
    \item Model Structure and Resolution Enhancement: Investigate the significance of model structural improvements and resolution enhancement in climate data processing. This includes assessing whether high-resolution models significantly outperform their low-resolution counterparts and the extent to which a mere increase in sample quantity can compensate for lower resolution in data analysis.
    
    \item Assessment of Climate Data Efficacy and Trend Representation: This study aims to scrutinize the current model's proficiency in capturing and representing ongoing climate change trends through its training on datasets of different mean states. It is still uncertain whether the multi-decadal training dataset diminishes the predictability of recent periods due to the impact of climate change. These adaptations will be aimed at augmenting the model's predictive capabilities for future climate change patterns and events.
\end{enumerate}

\section{Materials and Methods}

\subsection{Data Description}
The dataset employed in this research is composed of comprehensive global atmospheric fields sourced from the ECMWF Reanalysis v5 (ERA5) \cite{hersbach2020era5}. This extensive dataset is curated by the European Centre for Medium-Range Weather Forecasts (ECMWF), representing a significant reanalysis endeavor by the European forecasting authority. Within this dataset, the temporal snapshot encapsulates all 66 variables and is structured as a tensor with dimensions at 2.5-degree horizontal resolution (72 × 144 × 66). We use daily-mean data from the hourly data for each snapshot. For fair comparison to other existing models such as FourCastNet or Pangu-Weather, the training set encompasses a historical range from 1979 to 2015, inclusive, and the validation set is drawn from the subsequent two years, 2016 to 2017, ensuring that the model's performance is measured against recent data \cite{bi2023accurate}. Lastly, the testing set is designed to evaluate the model inference data from the initial conditions in 2018 same as the ECMWF operational forecasts (i.e., Every Monday and Thursday), providing a precise test of the model's predictive capability in out-of-sample scenarios.

\subsection{FourCastNet}
FourCastNet, which utilizes the Adaptive Fourier Neural Operator (AFNO) model, was developed by Guibas et al. This neural network integrates the Fourier Neural Operator (FNO) approach, which excels at modeling complex Partial Differential Equation (PDE) systems, with a robust Vision Transformer (ViT) backbone. The AFNO model excels in processing high-resolution data efficiently, contrasting sharply with traditional convolutional networks. For instance, the FourCastNet model's memory usage at a high resolution is just 10GB per batch, whereas a similarly adapted conventional 19-layer ResNet would require an impractical 83GB, highlighting AFNO's superior scalability. The FourCastNet processes input variables on a 720x1440 grid by projecting them onto a 2D grid of patches. These patches are then transformed into a sequence of multi-dimensional tokens, which, along with positional encodings, are processed through successive AFNO layers \cite{pathak2022fourcastnet}. 

\section{Modified Parts}

\subsection{Input Variables}
Our research model significantly expands upon the original FourCastNet framework by incorporating a more comprehensive set of variables. While the original FourCastNet model utilized a set of 20 variables, our model integrates a total of 66 variables, along with the addition of orography. This includes an extensive array of vertical level variables (u, v, t, q, z) across 12 atmospheric pressure levels, greatly surpassing the original model which covered vertical variables across fewer levels (1000 hPa, 850 hPa, 500 hPa, 50 hPa) \cite{betts2019near}. This substantial increase in the number of variables allows for a much more nuanced and detailed representation of the atmospheric conditions. The modified FourCastNet is trained to predict the same global 66-variable dataset on the next day using these input variables.

\begin{table}[h!]
	\caption{Comprehensive List of Key Atmospheric Variables for Climate Modeling, Corresponding Short Names, Vertical Measurement Levels hPa), and Units.}
	\centering
	\begin{tabularx}{\textwidth}{@{}lXXXX@{}}
		\toprule
		Variable name & Short name & Vertical levels (hPa) & Units & \\
		\midrule
		Zonal wind & U & 1000, 925, 850, 800, 700, 600, 500, 400, 300, 200, 100, 50 & m/s & \\
		\addlinespace
		Meridional wind & V & 1000, 925, 850, 800, 700, 600, 500, 400, 300, 200, 100, 50 & m/s & \\
		\addlinespace
		Temperature & T & 1000, 925, 850, 800, 700, 600, 500, 400, 300, 200, 100, 50 & K & \\
		\addlinespace
		Specific humidity & Q & 1000, 925, 850, 800, 700, 600, 500, 400, 300, 200, 100, 50 & kg/kg & \\
		\addlinespace
		Geopotential & Z & 1000, 925, 850, 800, 700, 600, 500, 400, 300, 200, 100, 50 & m\(^2\)/s\(^2\) & \\
		\addlinespace
		2m temperature & T2m & & K & \\
		\addlinespace
		Mean sea level pressure & MSL & & Pa & \\
		\addlinespace
		Surface air pressure & SP & & Pa & \\
		\addlinespace
		Total column vertically-integrated water vapour & TCWV & & kg/m\(^2\) & \\
		\addlinespace
		Skin temperature & SKT & & K & \\
		\addlinespace
		TOA incident solar radiation & TISR & & J/m\(^2\) & \\
		\bottomrule
	\end{tabularx}
	\label{tab:variables}
\end{table}

\subsection{Patch Size}
Initially, the patch size was firmly reduced to 1, following in the footsteps of Guo et al., who had previously downsized their model’s patch size to a precise 4x4 dimension \cite{guo2024fourcastnext}. They articulated a distinct divergence from conventional image classification tasks that utilize Imagenet data. In the context of FourCastNet, which operates by generating predictions at the same resolution as its input imagery, there is an inherent necessity to intricately capture and interpret detailed features. This requirement led to their strategic decision to diminish the patch size. In the sphere of our research, we engaged with data of a lower resolution relative to the original model. This adaptation necessitates a further refinement in the patch size, ensuring our approach remains optimally tailored to the specificities of our data's resolution. 

In our lower-resolution setting, the smaller patches enable a more detailed analysis of each image segment, leading to a more nuanced understanding of the data. By focusing on smaller sections of the image, our model becomes adept at identifying and interpreting key features that larger patches might overlook or misinterpret. Bhojanapalli et al.'s assertion underpins our methodology, emphasizing the importance of patch size in both resolution adaptation and adversarial robustness. It suggests that the capacity to maintain performance in the face of challenging conditions, such as lower resolutions or adversarial attacks, is significantly enhanced by choosing an appropriate patch size. In essence, the smaller patch size not only suits our specific use case of lower-resolution imagery but also fortifies the model against potential spatial manipulations, ensuring consistent and reliable performance across varying conditions \cite{bhojanapalli2021understanding}.

\subsection{Data Processing}
The original FourCastNet source code assumes that each file, representing a year's data, contains the same number of samples. This presumption overlooks one additional day of leap years, and such neglect could lead to inaccuracies in the total number of samples. To address this, we have modified the code to identify a specific year and subsequently ascertain its leap year status. The refined algorithm precisely assigns a count of 366 days for leap years, thereby aligning the sample count with the actual temporal span of the dataset

\begin{algorithm}
\caption{\textbf{Algorithm 1:} \textsc{Leap Year Detection}}
\begin{algorithmic}[1]
\Procedure{DetermineMaxSamples}{$year\_idx$}
    \State $year \gets \Call{year\_from\_idx}{year\_idx}$
    \If{$\Call{is\_leap\_year}{year}$}
        \State $max\_samples\_current\_year \gets 366$
        \State \textbf{print} "[DEBUG] Year $year$ is detected as a leap year."
    \Else
        \State $max\_samples\_current\_year \gets 365$
    \EndIf
\EndProcedure
\end{algorithmic}
\end{algorithm}

Moreover, the initial algorithm is designed such that upon encountering the final sample within a given year, it strategically resets the step variable to zero, thereby initializing the prediction process with the identity value. Therefore, to maintain the chronological integrity of our dataset, we also modified the original algorithm. The modified one is precisely calibrated to identify and exclude the final data point in a sequence which, crucially, lacks a subsequent target for prediction. This change ensures that every sample included in our model's training and validation phases has a valid and existent target. 

\begin{algorithm}
\caption{\textbf{Algorithm 2:} \textsc{Handling Last Index in Data Series}}
\begin{algorithmic}[1]
\Procedure{CheckSampleIndexValidity}{%
\par local\_idx, max\_samples\_current\_year,%
\par dt, year\_idx, n\_years, global\_idx%
}
    \If{$local\_idx \geq (max\_samples\_current\_year - dt)$}
        \If{$year\_idx = (n\_years - 1)$}
            \State \textbf{print} "[DEBUG] Skipping last sample from the last .nc file: global\_idx= $global\_idx$, local\_idx= $local\_idx$"
            \State \textbf{raise} IndexError "Skipping last sample from the last .nc file: global\_idx= $global\_idx$, local\_idx= $local\_idx$"
        \EndIf
    \EndIf
\EndProcedure
\end{algorithmic}
\end{algorithm}

\subsection{Data Augmentation}
We put forward a novel approach by utilizing a dataset with a time lag. This dataset includes a comprehensive series of daily-mean datasets, except for using different sliding 24-hour windows for average. It is defined that the conventional daily-mean averaging from 00:00 to 23:00 as ‘lag0’. Additionally, perturbed daily-mean data are also produced by the time-sliding method at a 6-hour interval lag, which uses different 24-hour sliding windows for daily-mean. For instance, a file representing a 6-hour lag would integrate data spanning from 06:00 to 23:00 of the preceding day, coupled with data from 00:00 to 05:00 of the subsequent day (i.e. lag6). Using this method, we produce additional daily-mean datasets by lagging 12-hour (lag12) and 18-hour (lag18). These datasets provide four times larger training samples for 1979-2015, thus offering a detailed and time-sensitive snapshot essential for our predictive analysis \cite{ho2021development}. We trained the model with 66 variables of the original daily-mean time series and that with the four times larger sample, called ‘lag0’ and ‘lag4x’respectively. From the trained model for one-day prediction, we performed auto-regressive inference until day 7.
\section{Result}
\subsection{Lagged Data}
For the t2m variable, the model trained with lag4x data results in notably lower Root Mean Squared Error (RMSE) values than that with lag0 data, particularly in the initial forecasting period. On day 1, the RMSE for lag4x is 0.488, compared to 0.588 for lag0 (Fig. 1). This trend continues on day 2 and day 3, with lag4x maintaining a lower RMSE. Similarly, the Anomaly Coefficient Correlation (ACC) score of lag4x data on day 1 is 0.924 for lag4x against 0.886 for lag0, indicating a clearer advantage for the model utilizing lagged data (Fig. 2).

The inference result of the z500 variable exhibits a similar pattern. The RMSE for lag4x var66 on day 1 stands at 48.23 $m^2s^{-2}$, considerably lower than the 61.16 for lag0. The ACC is also better for lag4x, with 0.989 on day 1, compared to 0.979 for lag0. In essence, the trained model with four times lagged data (lag4x) consistently outperforms the model with lag0 data across both variables and metrics. This disparity illustrates the enhanced predictive capabilities of models that incorporate more historical data points, indicating the current use of reanalysis samples is still insufficient for the DDWP framework. By utilizing lagged data, the model can better understand and predict the dynamics of the meteorological elements, leading to more accurate and reliable forecasts.

\begin{figure}[h]
    \centering
    \includegraphics[width=1\textwidth]{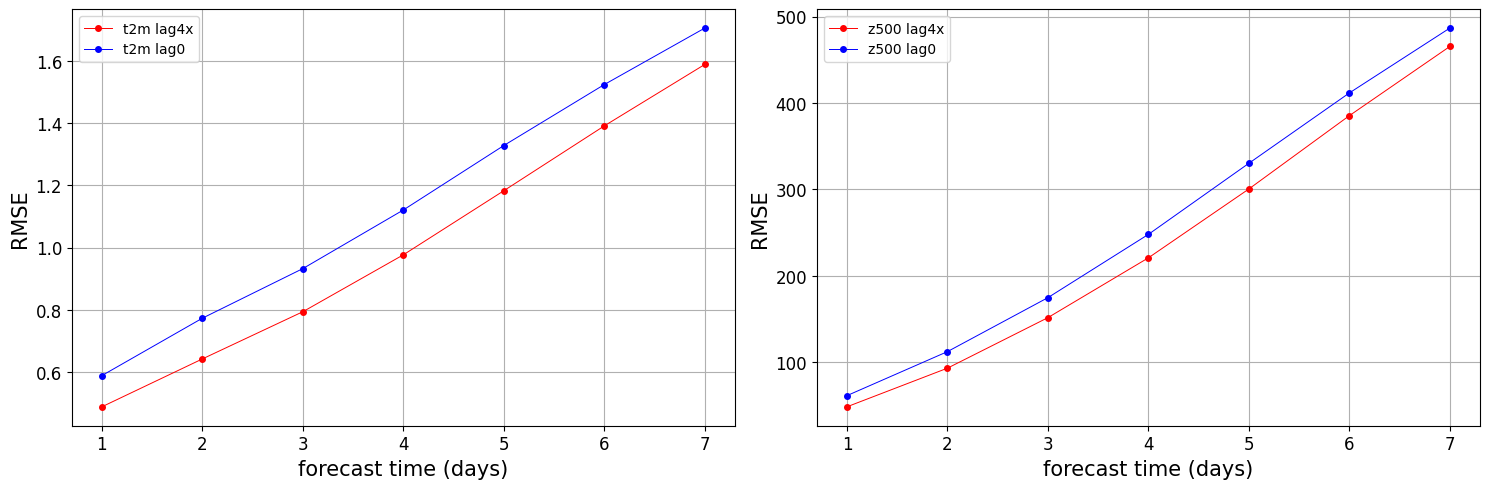}
    \caption{Comparative Analysis of Forecast Accuracy Over Forecast Time for 2m Temperature (t2m; K) and Geopotential at 500 hPa (\(z500; m^2s^{-2}\)) Using Lagged Data Inputs (RMSE).}
    \label{fig:fig1}
\end{figure}

\clearpage

\begin{figure}[h]
    \centering
    \includegraphics[width=1\textwidth]{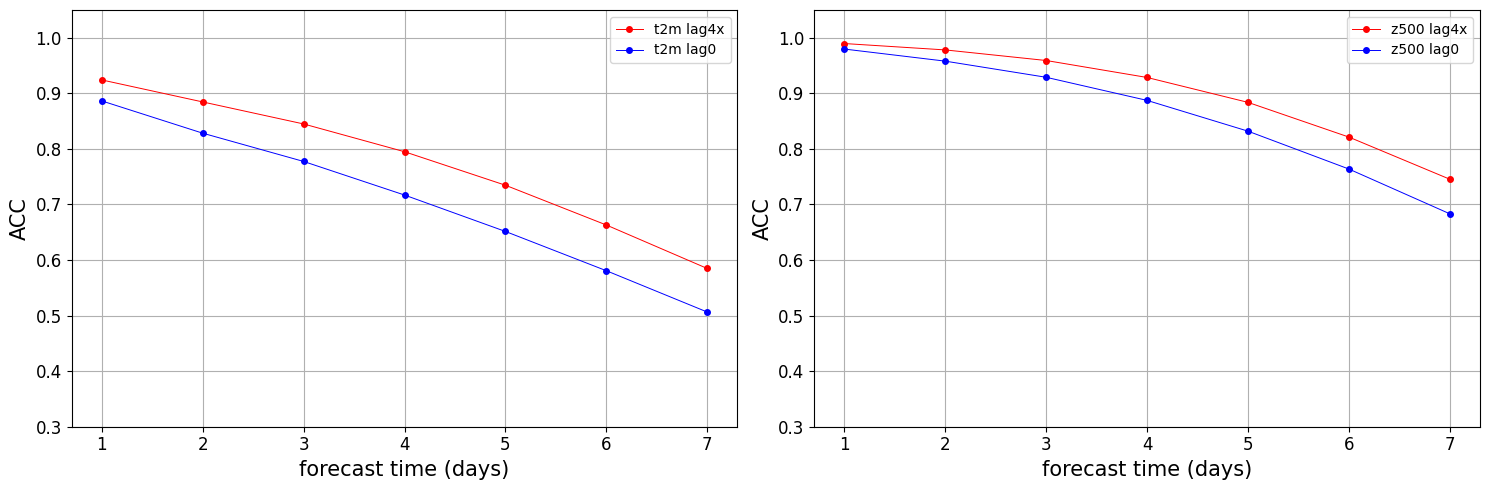}
    \caption{Comparative Analysis of ACC score Over Forecast Time for 2m Temperature (t2m) and Geopotential at 500 hPa (z500) Using Lagged Data Inputs.}
    \label{fig:fig2}
\end{figure}

\subsection{Comparative Analysis of Our Model and Existing Approaches}
In the comparative analysis, an array of four distinct models were employed: ClimaX, FourCastNet, Pangu-Weather, and GraphCast \cite{bi2023accurate}\cite{pathak2022fourcastnet}\cite{lam2022graphcast}\cite{nguyen2023climax}. The evaluation metrics, specifically RMSE and ACC, were extracted from the findings presented in the ClimaX study. Our model is based on daily mean data, while the others use hourly snapshots. This inconsistency could affect the score partly on the score of t2m, but it is less impactful on that of z500 because of its small diurnal variability (not shown). At a one-day forecast lead time of z500, our model achieves an RMSE value of 48.23 $m^2s^{-2}$, indicating that even at lower resolutions, the performance is comparable to that of Pangu-Weather. Moving forward to a three-day lead time, our model's prediction increases to 151.46 $m^2s^{-2}$. which also maintains our model's position as a competitive alternative. For the five-day forecast, our model yields a value of 300.41 $m^2s^{-2}$, which is in close competition with other models like GraphCast and Pangu-Weather. At the seven-day forecast, our model achieves a value of 465.39 $m^2s^{-2}$, which indicates a reliable extended forecast result compared to the other models. 

The results displayed in Table 2 provide a clear comparative analysis of our proposed model against other contemporary models in the domain of near-surface air temperature (t2m) prediction, with the time step indicating days ahead. At a one-day forecast, our model demonstrates a promising start with a value of 0.48 K, which is competitive with other models such as Pangu-Weather and GraphCast. Progressing to a three-day forecast, the value increases to 0.79 K, showcasing our model's adeptness at handling slightly longer-term predictions. This trend of increasing RMSE is maintained as we extend the forecast to five days, with our model achieving a value of 1.18 K. Our model records a value of 1.58 K in the seven-day forecast, which is competitive within the models. 

Overall, these results underscore the effectiveness of our proposed model in predicting t2m and z500 variables across various time steps, and they highlight the potential for our approach to contribute valuable insights in the realm of meteorological forecasting.

\begin{table}[h!]
	\caption{Comparative RMSE Values of Different Forecast Models for 2m Temperature (t2m) Prediction at Various Forecast Timesteps}
	\centering
	\begin{tabularx}{\textwidth}{@{}X X X X X X@{}}
		\toprule
		Timestep & ClimaX(5.625) & FCN(0.25) & Pangu & GraphCast & FCN\_lag4x \\
		\midrule
		24 hours (day1) & 1.00 & 0.95 & 0.72 & 0.62 & 0.48 \\
		\addlinespace 
		72 hours (day3) & 1.43 & 1.38 & 1.05 & 0.94 & 0.79 \\
		\addlinespace
		120 hours (day5) & 1.83 & 1.99 & 1.53 & 1.36 & 1.18 \\
		\addlinespace
		168 hours (day7) & 2.18 & 2.54 & 2.06 & 1.88 & 1.58 \\
		\bottomrule
	\end{tabularx}
	\label{tab:tabel2}
\end{table}

\clearpage

\begin{table}[h!]
	\caption{Comparative RMSE Values of Different Forecast Models for Geopotential at 500 hPa (z500) Prediction at Various Forecast Timesteps}
	\centering
	\begin{tabularx}{\textwidth}{@{}lXXXXX@{}}
		\toprule
		Timestep & ClimaX(5.625) & FCN(0.25) & Pangu & GraphCast & FCN\_lag4x \\
		\midrule
		24 hours (day1) & 96.19 & 81.31 & 42.23 & 38.77 & 48.23 \\
		\addlinespace
		72 hours (day3) & 244.08 & 251.96 & 133.12 & 125.78 & 151.46 \\
		\addlinespace
		120 hours (day5) & 440.4 & 483.44 & 295.63 & 271.65 & 300.41 \\
		\addlinespace
		168 hours (day7) & 599.43 & 680 & 504.9 & 466.53 & 465.39 \\
		\bottomrule
	\end{tabularx}
	\label{tab:table3}
\end{table}

\subsection{Impact of Recent Climate Data on Forecast Accuracy}
The experiment was conducted to test the hypothesis that climate data from the recent decade of 2006-2015 might provide a more accurate prediction basis than data from the earlier period of 1980-1989. This was predicated on the assumption that more contemporary data would be more representative of current climatic conditions and trends, potentially leading to enhanced forecasting accuracy. To address potential data scarcity for model training and to enhance the robustness of our predictive models, we strategically utilized the lag4x dataset (i.e., a four times larger dataset with lagged daily-means—lag0, lag6, lag12, and lag18). Upon evaluating the results of the test year of 2018, it becomes clear that utilizing the dataset from 2006-2015 markedly improves the accuracy of predictions for the z500 and t2m variables compared to using the dataset from 1980-1989. In the case of the z500 variable, the RMSE on the first forecast day is significantly lower with the 2006-2015 data (60.67 $m^2s^{-2}$) than with the 1980-1989 data (80.92 $m^2s^{-2}$), indicating 25\% reduction of forecast error using the training dataset in the recent decade. This pattern of greater accuracy is consistent on the third day (178.15 $m^2s^{-2}$ vs. 224.61 $m^2s^{-2}$), the fifth day (326.03 $m^2s^{-2}$ vs. 369.28 $m^2s^{-2}$), and extends to the seventh day (443.33 $m^2s^{-2}$ vs. 474.75 $m^2s^{-2}$). For the t2m variable, responsible for indicating near-surface temperature, the recent dataset also outperforms the older one. On the initial forecast day, the RMSE is tighter (0.64 K for 2006-2015 against 0.72 K for 1980-1989), suggesting a more accurate model. On the third and seventh days, the RMSE figures of 1.05 K and 1.90 K respectively for the recent dataset are superior to those of the older dataset (1.15 K and 2.31 K), illustrating the same trend of improved forecast precision. These comparisons highlight the robustness of the 2006-2015 dataset in providing a solid basis for more reliable and precise atmospheric predictions over one to seven days. However, the model trained with entire period data (1979-2015) shows a better skill than that with the recent decade (2006-2015), indicating that the increasing total sample size is more important than the impact of climate change within the dataset.

\begin{figure}[h!]
	\centering
	\includegraphics[width=\textwidth]{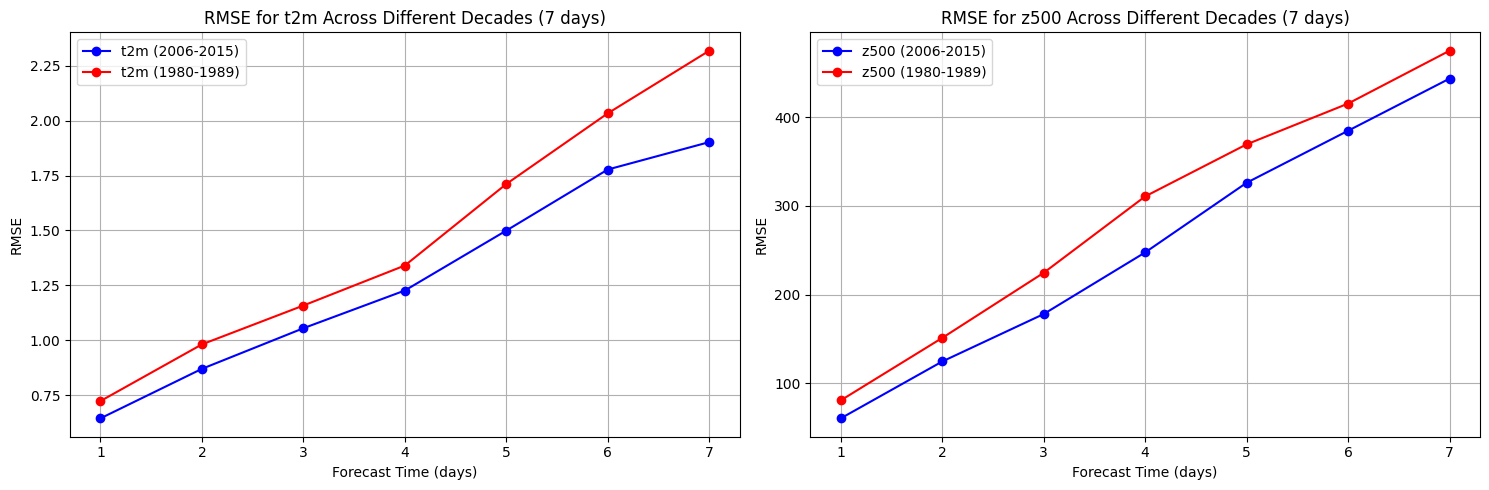}
	\caption{Impact of Decadal Climate Data on RMSE for 2m Temperature (t2m) and Geopotential at 500 hPa (z500) Forecasts.}
	\label{fig:fig3}
\end{figure}

\section{Conclusion}
This study has successfully met its primary objectives within the framework of data-driven weather prediction (DDWP), making significant contributions to the field of climate data analysis. The key findings of this research can be summarized as follows:

The research provides clear evidence that model structural improvements substantially impact climate data processing. While high-resolution models generally outperformed low-resolution ones, our study demonstrated that even lower-resolution data (2.5 degrees) could achieve reasonable prediction results. The more efficient training cost of a lower-resolution dataset can be beneficial to the case that the high-resolution is not necessary, such as subseasonal-to-seasonal (S2S) prediction with multi-ensemble. This finding is particularly relevant for scenarios where high-resolution data might not be readily available, suggesting that careful model structuring can partly compensate for resolution limitations.

Our research introduces the use of lagged data for data augmentation, which significantly enhanced the performance of our models through a quadrupled increase in sample size. This approach underscores that the multi-decadal reanalysis dataset used in the current DDWP models is still insufficient to accurately incorporate the complex atmospheric variability. Furthermore, the findings emphasize the significance of the recent training dataset for more accurate predictions, which is relatively less priority than the total sample size for training.

Additionally, our analysis confirms the current model's effectiveness in capturing and representing ongoing climate change trends. The model's predictive accuracy improves when the training dataset's background climate closely matches the test period's conditions, implying that the limited sample of data in the current climate condition is a hurdle to overcome for better DDWP skills. It is essential to expand training samples to encompass both present and forthcoming climate conditions to enhance the model's training efficacy for operational forecasting,

In conclusion, this study reaffirms the effectiveness of the DDWP framework in climate data analysis, while also highlighting areas for future enhancement. The insights gained from this research not only contribute to the field of climatology but also provide a solid foundation for further advancements in data-driven climate modeling and prediction.

\bibliographystyle{unsrtnat}
\bibliography{references}

\begin{thebibliography}{11}
\providecommand{\natexlab}[1]{#1}
\providecommand{\url}[1]{\texttt{#1}}
\expandafter\ifx\csname urlstyle\endcsname\relax
  \providecommand{\doi}[1]{doi: #1}\else
  \providecommand{\doi}{doi: \begingroup \urlstyle{rm}\Url}\fi

\bibitem[Dueben and Bauer(2018)]{dueben2018challenges}
P.~Dueben and P.~Bauer.
\newblock Challenges and design choices for global weather and climate models based on machine learning.
\newblock \emph{Geoscientific Model Development}, 11\penalty0 (10):\penalty0 3999--4009, 2018.
\newblock \doi{10.5194/gmd-11-3999-2018}.

\bibitem[Bauer et~al.(2023)]{bauer2023deep}
P.~Bauer et~al.
\newblock Deep learning and a changing economy in weather and climate prediction.
\newblock \emph{Nature Reviews Earth \& Environment}, 4\penalty0 (8):\penalty0 507--509, 2023.
\newblock \doi{10.1038/s43017-023-00468-z}.

\bibitem[Guo et~al.(2024)]{guo2024fourcastnext}
E.~Guo et~al.
\newblock Fourcastnext: Improving fourcastnet training with limited compute.
\newblock \emph{arXiv preprint arXiv:2401.05584}, 2024.

\bibitem[Hersbach et~al.(2020)]{hersbach2020era5}
H.~Hersbach et~al.
\newblock The era5 global reanalysis.
\newblock \emph{Quarterly Journal of the Royal Meteorological Society}, 146\penalty0 (730):\penalty0 1999--2049, 2020.

\bibitem[Bi et~al.(2023)]{bi2023accurate}
K.~Bi et~al.
\newblock Accurate medium-range global weather forecasting with 3d neural networks.
\newblock \emph{Nature}, 619\penalty0 (7970):\penalty0 533--538, 2023.
\newblock \doi{10.1038/s41586-023-06185-3}.

\bibitem[Pathak et~al.(2022)]{pathak2022fourcastnet}
J.~Pathak et~al.
\newblock Fourcastnet: A global data-driven high-resolution weather model using adaptive fourier neural operators.
\newblock \emph{arXiv preprint arXiv:2202.11214}, 2022.

\bibitem[Betts et~al.(2019)]{betts2019near}
A.~K. Betts et~al.
\newblock Near-surface biases in era5 over the canadian prairies.
\newblock \emph{Frontiers in Environmental Science}, 7:\penalty0 129, 2019.

\bibitem[Bhojanapalli et~al.(2021)]{bhojanapalli2021understanding}
S.~Bhojanapalli et~al.
\newblock Understanding robustness of transformers for image classification.
\newblock In \emph{Proceedings of the IEEE/CVF International Conference on Computer Vision}, pages 10231--10241, 2021.

\bibitem[Ho et~al.(2021)]{ho2021development}
C.~S. Ho et~al.
\newblock Development of a pm2.5 prediction model using a recurrent neural network algorithm for the seoul metropolitan area, republic of korea.
\newblock \emph{Atmospheric Environment}, 245:\penalty0 118021, 2021.
\newblock \doi{10.1016/j.atmosenv.2020.118021}.

\bibitem[Lam et~al.(2022)]{lam2022graphcast}
R.~Lam et~al.
\newblock Graphcast: Learning skillful medium-range global weather forecasting.
\newblock \emph{arXiv preprint arXiv:2212.12794}, 2022.

\bibitem[Nguyen et~al.(2023)]{nguyen2023climax}
T.~Nguyen et~al.
\newblock Climax: A foundation model for weather and climate.
\newblock \emph{arXiv preprint arXiv:2301.10343}, 2023.

\end{thebibliography}

\end{document}